\documentclass[letterpaper, 10 pt, conference]{ieeeconf}
\IEEEoverridecommandlockouts

\usepackage{cite}
\usepackage{amsmath,amssymb,amsfonts}
\usepackage{dsfont}
\usepackage{mathtools}
\usepackage{algorithm,algpseudocode}
\usepackage{graphicx}
\usepackage[caption=false]{subfig}
\usepackage{textcomp}
\usepackage{xcolor}
\usepackage{flushend}
\usepackage[hidelinks]{hyperref} 
\def\BibTeX{{\rm B\kern-.05em{\sc i\kern-.025em b}\kern-.08em
    T\kern-.1667em\lower.7ex\hbox{E}\kern-.125emX}}

\DeclareMathOperator{\vect}{vec}

\newcommand{\Bezier}{B\'{e}zier }

\floatname{algorithm}{Algorithm}

\DeclareMathOperator*{\argmin}{arg\,min}

\title{\LARGE \bf Local Parametric Surface Approximation With \\ Automatic Order Selection From Position Data}

\author{Michael R. Walker II, \IEEEmembership{Member, IEEE}
	\thanks{This work was supported by Stereotaxis, Inc.}
	\thanks{M. R. Walker II is with Stereotaxis, Inc., 4320 Forest Park Ave., St. Louis, Missouri, 63108, USA
		{\tt\small mwalkerii@wustl.edu}}%
	\thanks{Source code and example data are available:  \href{https://github.com/mrw2ee/BezBic}{github.com/mrw2ee/BezBic}}
}

\begin{document}


\maketitle
\thispagestyle{empty}
\pagestyle{empty}

\begin{abstract}
Acquiring an anatomical map from position data is important for medical applications where catheters interact with soft tissues. To improve autonomous navigation in these settings, we require information beyond nonparametric maps typically available. We present an algorithm for local surface approximation from position data with automatic surface order selection. The traditional surface fitting objective function is derived from a Bayesian perspective. Posterior probabilities from the occupancy map are incorporated as weights on points selected for surface fitting. Our novel iterative algorithm incorporates surface order selection using the Bayesian information criterion. Simulations demonstrate the ability to automatically select surface order consistent with the latent surface in the presence of noise. Results on human procedure data are also presented.

\end{abstract}


\section{Introduction}
\begin{figure}
	\centering
	\includegraphics[width=\hsize]{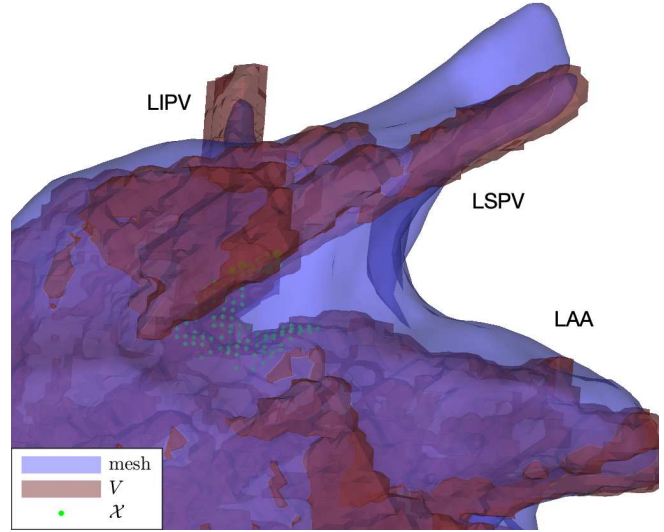}
	\caption{Anatomic mapping data in left atrium. The surface presented to the physician for navigation is labeled {\it mesh}. A level set of the occupancy map is labeled $V$. Additionally, we indicate a local selection of boundary voxels $\mathcal{X}$. The left inferior pulmonary vein (LIPV), left superior pulmonary vein (LSPV), and left atrial appendage (LAA) are labeled. The voxels indicated by $\mathcal{X}$ are on the ridge between the pulmonary veins and the LAA. For size reference, voxels are 1mm\textsuperscript{3}.}\label{fig:leftatrium}
\end{figure}

Anatomic maps used by human navigators exhibit significant interpolation when contrasted against occupancy maps aggregating smoothed position data (see Figure \ref{fig:leftatrium}). In cardiac catheter ablation surgery, the occupancy map's utility extends beyond traditional autonomous navigation tasks (e.g. path planning, obstacle avoidance) as catheter-tissue interactions affect catheter response\cite{Tunay2013}. Here we present a robust algorithm for local parametric surface approximation providing new information from noisy, incomplete data in real time (1Hz updates).

This paper has three main contributions. First we describe a novel algorithm for point selection from an occupancy map for non-planar surfaces. Second, we augment the traditional objective function for surface fitting to include posterior probabilities available from occupancy maps. Third, and most significantly, we incorporate automatic surface order selection in the iterative minimization algorithm, which is critical for distinguishing curvature of the latent surface from noise (errors) in the data.

\begin{figure}
	\centering
	\def\svgwidth{\hsize}
	\begin{footnotesize}
		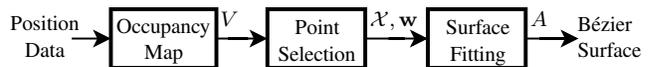
	\end{footnotesize}
	\caption{Process for local surface approximation.}\label{fig:process}
\end{figure}

Here we approximate anatomic surfaces using position data, from the therapeutic catheter, alone. The process is outlined in Figure \ref{fig:process}. Position data are typically available with sub-mm precision \cite{Bourier2014}. These data are subject to strong heartbeat and respiratory motion \cite{McClelland2013} which we suppress using an unpublished algorithm\cite{Walker2016b}. Since localization data is available, our problem does not include simultaneous localization and mapping (SLAM)\cite{Cadena2016}, and constructing an occupancy map \cite{Hornung2013} is straight forward. Posterior probabilities can be assigned to voxels based on additional sensor data (e.g. tip force or magnetic torque). From the occupancy map, we consider a dense binary matrix, $V\in\left\{0,1\right\}^{n\times m\times p}$, thresholding the known interior volume. From this, we identify an unorganized collection of points, ${\mathcal{X} = \left\{\mathbf{x}_i \in \mathbb{R}^3\right\}}$, which we interpret as a noisy, non-uniform sampling of the latent surface. From these points, we determine a set of control points, ${A\in\mathbb{R}^{\left(n_u+1\right)\times\left(n_v+1\right)\times 3}}$, defining a \Bezier surface. Parameters $n_u,n_v\in\mathbb{N}_{>0}$ set the surface order and are determined automatically.

There is considerable published work on the fitting of \Bezier curves and related generalizations to point clouds. At a high level, we distinguish gradient-free search methods \cite{Galvez2013, Iglesias2016} from gradient-based search methods (e.g. Gauss-Newton) \cite{Pastva1998,Pottmann2002,Wang2006,Liu2008} (among others). In most cases, the surface order is assumed fixed. We find the work of Iglesias et al. an interesting exception for their inclusion of the Bayesian information criterion (BIC) for surface order selection \cite{Iglesias2016}. Our approach to surface fitting can be summarized as alternating updates of the location parameters and control points using the distance minimization method (PDM) (see \cite{Wang2006} for broader context). Our approach is unique in that we utilize the BIC to conditionally increase the surface order at each update of the control points. Once the surface order is fixed, other methods have demonstrated faster convergence rates\cite{Zheng2012, Bo2012} and could be used to refine surface approximations more efficiently if necessary.

The rest of this paper is organized as follows. In Section \ref{sec:Model} we present analytic and statistical models guiding algorithm design. In Section \ref{sec:Algorithms} we describe algorithms for point selection, model fitting, and model selection. In Section \ref{sec:Results} we quantify performance using analytic simulations and demonstrate results on human procedure data. Final remarks are given in Section \ref{sec:Conclusion}.

\section{Model} \label{sec:Model}
This work is focused on fitting a parametric surface to available data. To this end we describe two models: an analytic model for the surface, and a stochastic model for the data. We will subsequently describe surface fitting as maximum-likelihood estimation, and the same stochastic models will be used to determine surface order automatically.

\subsection{Parametric surface}
We choose \Bezier surfaces for our analytic model based on their broad adoption and efficient use of parameters. Fundamental to the definition of a \Bezier surface are the Bernstein polynomials $b: \mathbb{R}^1\rightarrow\mathbb{R}^1$
\begin{equation}
b(u;i,n) = \begin{pmatrix} n\\i \end{pmatrix}u^i(1-u)^{n-i},\quad\, \forall i\in[0,\ldots,n].\label{eq:bernsteinpoly}
\end{equation}
We collect the values of all basis functions as the vector-valued function $\mathbf{b}:\mathbb{R}\rightarrow\mathbb{R}^{n+1}$
\begin{equation}
\mathbf{b}(u;n) = \begin{bmatrix} b(u;0,n) & b(u;1,n) & \cdots & b(u;n,n) \end{bmatrix}^T.\label{eq:bbold}
\end{equation}
The parameter $n$ determines the length of $\mathbf{b}$, $n+1$, and we will omit it when clear from context. 

A \Bezier surface is simply a weighted combination of Bernstein polynomials. For each coordinate, we define $s : \mathbb{R}^2\rightarrow\mathbb{R}$
\begin{equation}
s(u,v;A_{**k}) = \mathbf{b}(u;n_u)^T A_{**k} \mathbf{b}(v;n_v)\label{eq:sscalar}.
\end{equation}
Here we use $n_u$ and $n_v$ to represent the polynomial order in directions $u$ and $v$, respectively. The function is parameterized by a portion of the matrix $A\in\mathbb{R}^{(n_u+1)\times (n_v+1)\times 3}$. We define the \Bezier surface, concatenating three copies of \eqref{eq:sscalar}, as a mapping $\mathbf{s}:\mathbb{R}^2\rightarrow \mathbb{R}^3$
\begin{equation}
\mathbf{s}(u,v;A) = \begin{bmatrix} s(u,v;A_{**1}) \\ s(u,v;A_{**2}) \\ s(u,v;A_{**3})\end{bmatrix}.\label{eq:svect}
\end{equation}
In subsequent expressions it will be convenient to reshape $A$ as a two-dimensional matrix ${\bar{A}\in\mathbb{R}^{(n_u+1) (n_v+1)\times 3}}$. We make this distinction as $n_u,n_v$ are clear from $A$, but ambiguous from $\bar{A}$ alone. We restate \eqref{eq:sscalar} and \eqref{eq:svect}
\begin{align}
s(u,v;A_{**k}) &= \vect\left( A_{**k} \right)^T\left(\mathbf{b}(v;n_v)\otimes\mathbf{b}(u;n_u)\right)\\
\mathbf{s}(u,v;A) &= \bar{A}^T\left(\mathbf{b}(v;n_v)\otimes\mathbf{b}(u;n_u)\right)\label{eq:skron}
\end{align}
using $\otimes$ to represent the Kronecker product.

\subsection{Stochastic Data}
Let $\mathcal{X} = \left\{\mathbf{x}_i\in\mathbb{R}^3 : i = 1,\ldots,n_x\right\}$ represent an indexed set of points. We model the data
\begin{equation}
\mathbf{x}_i = \mathbf{s}\left(u_i,v_i;A\right) + \mathbf{y}_i, \quad i\in\left\{1,\ldots,n_x\right\}.\label{eq:xmdl}
\end{equation}
Here $\mathbf{y}_i\in\mathbb{R}^3$ represents a stochastic error term. This offset is due to a number of factors including quantization of the volume, biological motion, and model error. Motivated by the central limit theorem, we consider the errors to be Gaussian distributed.  For simplicity, we model these terms as independent and distributed as ${\mathbf{y}_i \sim \mathcal{N}\left(0, I_3\,\sigma^2/w_i^2\right)}$.  The weights, $w_i\in\mathbb{R}_{>0}$, represent the posterior probabilities of the occupancy map encoding our confidence in each $x_i\in\mathcal{X}$. Reducing $w_i$, associated with $x_i$, increases the modeled variance of corresponding $y_i$. 

Let ${\theta = \left\{\mathbf{u},\mathbf{v},A,\sigma^2\right\}}$ collectively refer to the model parameters. We define column vectors $\mathbf{u}, \mathbf{v}\in\mathbb{R}^{n_x}$ comprising the coordinates $\left(u_i,v_i\right)$ associated with the indexed elements of $\mathcal{X}$.The log-likelihood of the data, parameterized by $\theta$, reads \cite{moon2000}
\begin{multline}
\ell (\mathcal{X};\theta) = -n_x\frac{3}{2}\ln\left(2\pi\right)\\
+\sum_{i=1}^{n_x} \left[-\frac{3}{2} \ln\left(\frac{\sigma^2}{w_i^2}\right) - \frac{1}{2}\frac{w_i^2}{\sigma^2}\left\|\mathbf{x}_i - \mathbf{s}\left(u_i,v_i;A\right)\right\|_{\ell 2}^2\right].\label{eq:ll}
\end{multline}
The maximum likelihood estimate of $\theta$ are the model parameters maximizing \eqref{eq:ll}. This expression will also prove useful in selecting $n_u$ and $n_v$ which determine the size of $A$.

\section{Algorithms}\label{sec:Algorithms}
For our problem, $\mathcal{X}$ are not immediately available. They must be determined based on a record of the known volume. In the following, we first present an algorithm for selecting $\mathcal{X}$, and $w_i$, from a binary occupancy map. Then, we fit the surface parameters $A$, $\mathbf{u}$, and $\mathbf{v}$ for fixed $n_u$, $n_v$. Finally, we consider selection of $n_u$, $n_v$.

\subsection{Point Selection}\label{sec:pointselect}
Let $V\in\left\{0,1\right\}^{n\times m\times p}$ indicate voxels visited by the catheter obtained by thresholding the occupancy map. Our objective in this section is to identify a subset of these voxels which locally approximate a single surface. We identify interior boundary voxels using convolution and thresholding
\begin{equation}
\widetilde{V} = \left(W*V\right) \geq \epsilon
\end{equation}
where $W$ is a discrete convolutional kernel matrix and $\epsilon$ is a scalar threshold. There is some flexibility in selecting $W$. We use the three-dimensional Laplacian kernel 
\begin{equation}
\left[W_\epsilon\right]_{i,j,k} = \begin{dcases*} 26 & $i=j=k=2$\\
-1 & otherwise\end{dcases*}
\end{equation}
such that the threshold specifies the minimum number of exterior voxels in a $3\times3\times3$ neighborhood. In application, we found $\epsilon=9$ provided reasonable balance of sensitivity and specificity for our problem.

The voxels indicated by $\widetilde{V}$ typically compose multiple surfaces. We attribute this to mislabeled voxels in matrix $V$. For example, some interior voxels may not be visited by the catheter. Next we seek to identify a local subset of voxels, indicated by $\widetilde{V}$, associated with a single surface.

We consider two points $(i,j,k)$, and $(i',j',k')$ associated with the same surface  when they are both indicated by $\widetilde{V}$ and are within the same neighborhood. We summarize these requirements
\begin{equation}
\left[\widetilde{V}\right]_{i,j,k} = \left[\widetilde{V}\right]_{i',j',k'} = \left[J_l*\delta_{i,j,k}\right]_{i',j',k'} = 1. \label{eq:connected}
\end{equation}
In the final equality, we use $\delta_{i,j,k}$ to represent the indicator matrix where the element $(i,j,k)$ is one. We use $J_l$ to indicate the $l\times l\times l$ matrix of all ones. Starting with an initial point $(i,j,k)$ such that $\left[\widetilde{V}\right]_{i,j,k}=1$, we expand the collection of points iteratively convolving with $J$. At each iteration, the iteration number approximates the distance along the surface from newly added points to the initial point. The point selection process is summarized in Algorithm \ref{alg:pointselect}.

\begin{algorithm}
	\caption{\textsc{PointSelection}. Identify points locally approximating a single surface about a query point. Input binary matrix $V$ indicates interior voxels, and $\delta$ indicates the query point. We assume the size of kernels $W,J$ do not exceed the size of $V$. The indicated points are associated with nonzero entries in $R$. The values of $R$ approximate an inverse distance, along the surface, to the query point. We use $*$ to indicate 3D convolution and $\circ$ to indicate element-wise multiplication.}
	\label{alg:pointselect}
	\algloopdefx{Find}[1][\textbf{find}]{#1}
	\begin{algorithmic}[1]
		\Require $V,\delta\in\left\{0,1\right\}^{n\times m\times p}$; $W,J\in\mathbb{Z}^{n'\times m'\times p'}$
		\Ensure  $R\in\mathbb{N}_{\geq 0}^{n\times m\times p}$
		\State $\widetilde{V} \gets \left(W*V\right) > 0$
		\State $R \gets \left(J*\delta\right)\circ \widetilde{V}$
		\For{$t = 1,2,\ldots$}
		\State $R \gets R +  \left(J*R\right)\circ \widetilde{V}$
		\EndFor
	\end{algorithmic}
\end{algorithm}

In general, we assume Algorithm \ref{alg:pointselect} is applied locally such that 3D convolutions are performed quickly. When $V$ represents a truncated set of the known volume, voxels along the perimeter of $\widetilde{V}$ may be mislabeled. A maximum number of iterations should be enforced such that $J*R$ never indicates perimeter voxels.

In remaining algorithms, we will not reference $V$ or $R$. Rather, we use $R$ to determine an indexed set of points $\mathcal{X} = \left\{\mathbf{x}_i\in\mathbb{R}^3 : i = 1,\ldots,n_x\right\}$. Each element, $\mathbf{x}_i\in \mathcal{X}$, is associated with a nonzero entry of $R$. Additionally, we account for an indexed set of weights $w_i>0$.  These could be used to emphasize points closer to the query point, or to represent stochastic priors from an occupancy map \cite{Hornung2013}.

\subsection{Model Fitting}\label{sec:modelfit}
Only $\mathbf{s}$ in \eqref{eq:ll} is affected by the parameters $A$, $\mathbf{u}$, and $\mathbf{v}$.  The maximum likelihood estimates of these parameters are found minimizing 
\begin{equation}
f(A,\mathbf{u},\mathbf{v}) = \frac{1}{2}\sum_{i=1}^{n_x} w_i^2 \left\| \mathbf{x}_i - \mathbf{s}\left(u_i,v_i;A\right)\right\|_{\ell 2}^2.\label{eq:fauv}
\end{equation}
This expression is challenging to minimize since \eqref{eq:sscalar} involves a high-order product of its arguments. However, the arguments provide a natural decomposition for block coordinated descent \cite{Bertsekas2016}. We define
\begin{align}
\mathcal{P}1 &: &\mathbf{u}^{(t+1)},\mathbf{v}^{(t+1)} &= \argmin_{\bf{u},\bf{v}} f(A^{(t)},\bf{u},\bf{v})\label{eq:minuv}\\
\mathcal{P}2 &: &A^{(t+1)} &= \argmin_A f(A, \mathbf{u}^{(t+1)},\mathbf{v}^{(t+1)}) \label{eq:minA}
\end{align}
and iteratively update estimates of the control points and location parameters.

By fixing $A$, \eqref{eq:fauv} becomes separable. In this way, $\mathcal{P}1$ decomposes as $n_x$ two-dimensional optimization problems. For each $x_i\in\mathcal{X}$, we solve 
\begin{equation}
\hat{u}_i,\hat{v}_i = \arg\min_{u,v\in\mathbb{R}} g\left(u,v;\mathbf{x}_i,A\right)\label{eq:argminuv}
\end{equation}
where $g:\mathbb{R}^2\rightarrow\mathbb{R}^1$
\begin{equation}
g(u,v;\mathbf{x},A) \coloneqq \frac{1}{2}\left\| \mathbf{x} - \mathbf{s}\left(u,v;A\right)\right\|_{\ell 2}^2.\label{eq:g}
\end{equation}
The gradient and Hessian of \eqref{eq:g}, with respect to $u$ and $v$, are available analytically (see Appendix \ref{app:gradhess} for details). They are not constant with respect to $u$ and $v$, and the Hessian is not guaranteed to be positive definite. However, a local minimum can be identified quickly using Newton's method with Armijo backtracking \cite{Bertsekas2016}.

In contrast, $\mathcal{P}2$ has a closed-form solution. For convenience we horizontally concatenate $\mathbf{x}_n$ as $X\in\mathbb{R}^{3\times n_x}$, and define ${B\in\mathbb{R}^{ (n_u+1)(n_v+1)\times n_x}}$
\begin{equation}
B = \begin{bmatrix}  \mathbf{b}(\left[\mathbf{v}\right]_1) \otimes \mathbf{b}(\left[\mathbf{u}\right]_1) & \cdots &  \mathbf{b}(\left[\mathbf{v}\right]_{n_x}) \otimes \mathbf{b}(\left[\mathbf{u}\right]_{n_x}) \end{bmatrix}.\label{eq:B}
\end{equation}
Using $w_i$, we compose the diagonal matrix ${\Lambda\in \mathbb{R}_{>0}^{n_x\times n_x}}$. We then restate \eqref{eq:fauv} as a matrix equation
\begin{equation}
f(A,{\bf u},{\bf v}) = \frac{1}{2} \left\|\Lambda B^T \bar{A} - \Lambda X^T\right\|_{F}^2, \label{eq:fauvmatrix}
\end{equation}
replacing the vector norm with the matrix norm. In this form, the solution to $\mathcal{P}2$ is obvious.

Often $\mathcal{P}2$ is poorly scaled, and without regularization the optimal $A$ will include control points far from the elements of $\mathcal{X}$. Regularization has been addressed previously (e.g. Tikhonov \cite{Jing2005}, and the fairing term \cite{Wang2006, Bo2012}). Using Tikhonov regularization the solution remains available in closed form.  Without loss of generality, we assume $\mathcal{X}$ are centered about the origin. This assumption may require translation of the initial point cloud $\widetilde{\mathcal{X}}$ before fitting and translation of $A$ after fitting summarized as
\begin{equation*}
X = \widetilde{X} - \mathbf{x}_0 \mathds{1}^T  \text{, }\quad \widetilde{A} =  \bar{A}+ \mathds{1} \mathbf{x}_0^T.
\end{equation*}
Here we use $\mathds{1}$ to represent a vector of all ones. We define the regularized function
\begin{equation}
f_\lambda(A,{\bf u},{\bf v}) = \frac{1}{2} \left\|\Lambda B^T \bar{A} - \Lambda X^T\right\|_{F}^2 + \frac{\lambda}{2} \left\| \bar{A} \right\|_{F}^2.\label{eq:fauvreg}
\end{equation}
The $A$ minimizing \eqref{eq:fauvreg} is found solving the linear system of equations
\begin{equation}
\left(B\Lambda^2 B^T + \lambda I\right)\bar{A} =  B\Lambda^2 X^T.\label{eq:Aregularized}
\end{equation}

So far, we have assumed $n_u$ and $n_v$ constant. They determine the size of $A$ and $B$ but have no affect on the size of $\mathbf{u}$ or $\mathbf{v}$. In this way, each iteration of $\mathcal{P}2$ provides an opportunity to change $n_u$, $n_v$. Next, we consider selection of these parameters.

\subsection{Model Selection}
We can force \eqref{eq:ll} arbitrarily small by selecting large $n_u$ and $n_v$. However, increasing these parameters leads to higher order surfaces which are not anatomically accurate. We cast the problem of choosing $n_u$ and $n_v$ as model selection. For this we employ the Bayesian information criteria (BIC). 

Let $q$ index the candidate models $\theta_q$. In our case, each $\theta_q$ comprises the optimal parameters associated with a unique surface order $(n_u,n_v)$. We seek the model which most-likely generated the available data. As $n_x$ increases, the BIC asymptotically approximates the joint log likelihood of the data and model \cite{Lanterman2001}
\begin{align}
\ln p\left(\mathcal{X},q\right) \approx \ell (\mathcal{X};\theta_q) - \frac{d_q}{2}\ln n_x.\label{eq:bic}
\end{align}
In this approximation $\theta_q$ represents the maximum likelihood estimates for model $q$. The scalar $d_q$ indicates the total number of model parameters associated with the model $q$
\begin{equation}
d_q = 2 n_x + 3\left(n_u+1\right)\left(n_v+1\right) + 1.\label{eq:d}
\end{equation}
Expression \eqref{eq:bic} provides a regularized objective function for model order selection. Gains in $\ell (\mathcal{X};\theta_q)$ are offset by additional model parameters.

In \eqref{eq:bic} we  make use of \eqref{eq:ll} which requires $\sigma^2$. Given maximum-likelihood estimates $A$, $\mathbf{u}$, $\mathbf{v}$ (as detailed in Section \ref{sec:modelfit}), the maximum-likelihood estimate of $\sigma^2$ is then
\begin{equation}
\begin{split}
\hat{\sigma}^2 &= \frac{1}{3n_x} \sum_{i=1}^{n_x} w_i^2\left\|\mathbf{x}_i - \mathbf{s}_i\right\|_{\ell 2}^2\label{eq:sigma2}\\
&= \frac{2}{3 n_x} f\left(A,\mathbf{u},\mathbf{v}\right).
\end{split}
\end{equation}
Plugging \eqref{eq:sigma2} into \eqref{eq:ll}, we retain only terms of $\sigma_q$ and $d_q$ and define the statistic
\begin{equation}
t_{q} = -3 n_x \ln \sigma_q^2 - d_q \ln n_x.\label{eq:bic_abbrev}
\end{equation}
Here $q$ indexes candidate models $\theta_q$, which includes $\sigma_q^2$, and determines $d_q$ according to \eqref{eq:d}. The model selection maximizing \eqref{eq:bic} is equivalent to selecting $q$ maximizing the statistic $t_q$. Algorithm \ref{alg:MdlSelect} describes the process of estimating $A$ for multiple pairs $\left(n_u,n_v\right)$ and selecting the result yielding the largest $t$.

\begin{algorithm}
	\caption{\textsc{MdlSelect}. Estimate $A$ and statistic $t$ while increasing model order. Return the model with the largest statistic. Here \textsc{ComputeA} refers to the solution for \eqref{eq:Aregularized}, and \textsc{ComputeSigma2} refers to \eqref{eq:sigma2}. While the parameters $\mathbf{u},\mathbf{v}$ are returned in $\theta$, they are not changed.}
	\label{alg:MdlSelect}
	\algloopdefx{Find}[1][\textbf{find}]{#1}
	\begin{algorithmic}[1]
		\Require $X\in\mathbb{R}^{3\times n_x}; \mathbf{w}\in\mathbb{R}^{n_x}_{>0}; \mathbf{u},\mathbf{v}\in\mathbb{R}^{n_x}; n_u,n_v\in\mathbb{N}_{>0};$
		\hspace{\algorithmicindent}${\lambda \in\mathbb{R}_{\geq 0}}$
		\Ensure  $\theta$
		\State $q \gets 0$
		\For{$n_v' = n_u$ to $n_u+1$}
			\For{$n_v' = n_v$ to $n_v+1$}
			\State $q\gets q+1$
			\State $A \gets \textsc{ComputeA}\left(X,\mathbf{w},\mathbf{u},\mathbf{v},n_u',n_v',\lambda\right)$\Comment{\eqref{eq:Aregularized}}
			\State $\sigma^2 \gets \textsc{ComputeSigma2}\left(X,\mathbf{w},A, \mathbf{u},\mathbf{v}\right)$\Comment{\eqref{eq:sigma2}}
			\State $t_q \gets \textsc{ComputeT}\left(\theta\right)$\Comment{\eqref{eq:bic_abbrev}}
			\EndFor
		\EndFor
		\State $i \gets \arg\max_{i\in\left\{1,\,\ldots\,,\,q\right\}} t_i$ \Comment{Maximize BIC}
		\State $\theta\gets \theta_i$
	\end{algorithmic}
\end{algorithm}

The surface fitting process is given in Algorithm \eqref{alg:fitsurface}. We assume $\mathbf{u},\mathbf{v}$ have been initialized, for example, projecting $X$ onto a 2D subspace using SVD. With each update of $A$, in \textsc{MdlSelect}, we potentially increase surface order according to the BIC. Changes to $\mathbf{u},\mathbf{v}$, and $\sigma^2$ are all useful for stopping criteria (not addressed here).

\begin{algorithm}
	\caption{\textsc{FitSurface}. Fit a surface to a collection of points iteratively updating $\mathbf{u},\mathbf{v}$, and $A$.  \textsc{UpdatePoint} refers to \eqref{eq:argminuv}, which requires an iterative solver.}
	\label{alg:fitsurface}
	\algloopdefx{Find}[1][\textbf{find}]{#1}
	\begin{algorithmic}[1]
		\Require $X\in\mathbb{R}^{3\times n_x}; \mathbf{u},\mathbf{v}\in\mathbb{R}^{n_x}; \mathbf{w}\in\mathbb{R}^{n_x}_{>0};\lambda>0$
		\Ensure  $\theta$
		\State $n_u,n_v \gets 1$
		\State $\theta \gets \textsc{MdlSelect}\left(X,\mathbf{w},\mathbf{u},\mathbf{v},n_u,n_v,\lambda\right)$\Comment{Algorithm \ref{alg:MdlSelect}}
		\For{$t = 1,2, \ldots$}
		\State $A, n_u, n_v  \gets A(\theta), n_u(\theta), n_v(\theta)$ \Comment{Expand last $\theta$}
		\For{$i = 1$ to $n_x$}\Comment{Update $\mathbf{u}$, $\mathbf{v}$}
		\State $u_i,v_i \gets \textsc{UpdatePoint}\left(\mathbf{x}_i, A, u_i, v_i\right)$\Comment{\eqref{eq:argminuv} }
		\EndFor
		\State $\theta \gets \textsc{MdlSelect}\left(X,\mathbf{w},\mathbf{u},\mathbf{v},n_u,n_v,\lambda\right)$\Comment{Update $A$}
		\EndFor
	\end{algorithmic}
\end{algorithm}

The statistical interpretation of \eqref{eq:bic}, as an approximation for the joint log-likelihood, is somewhat disingenuous for our problem. The asymptotic approximation of the BIC is not accurate for our problem sizes ($n_x\sim 100$). Additionally, the criterion requires maximum likelihood estimates of the parameters which are not resolved during the iterative algorithm. The implication here is that $q$, maximizing $t_q$, may fluctuate while the algorithm converges. Additionally, switching $q$ online (as in Algorithm \eqref{alg:fitsurface}) may yield different results in contrast with the brute-force approach of solving for each model order independently and then applying model selection. However, the brute-force approach is computationally expensive. Here we use the right hand side of \eqref{eq:bic} as a regularized objective function guiding allocation of computational resources.

\section{Simulations and Results}\label{sec:Results}
To demonstrate the benefits of our approach, we quantify performance with simulations and present qualitative results on human procedure data. Simulations are primarily designed to demonstrate automatic surface order selection: avoiding both overfitting and underfitting the latent surface. 

\begin{figure}
	\centering
	\includegraphics[width=\hsize]{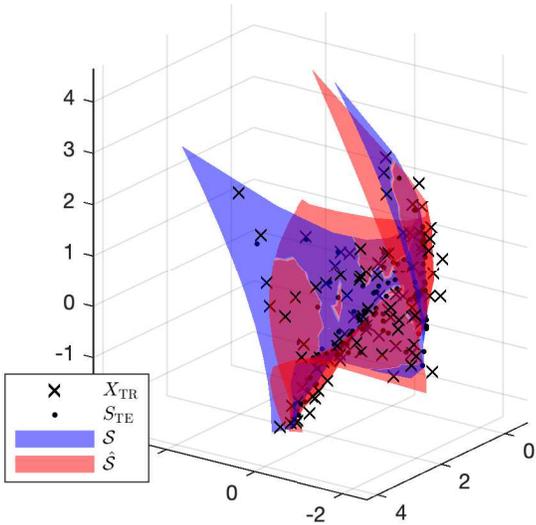}
	\caption{Visualization of surface approximation. The training and testing data are indicated by $X_\textrm{TR}$ and $S_\textrm{TE}$, respectively. The reference surface and approximation are indicated by $\mathcal{S}$ and $\hat{\mathcal{S}}$, respectively. Here $n_\textrm{TR}=100$, and $\sigma_y^2 = 0.02$.}\label{fig:rosenbrock}
\end{figure}

We simulate a notional reference surface as an infinite collection of points $\mathcal{S}\subset\mathbb{R}^3$. We then define two unique subsets, $\mathcal{S}_\textrm{TR}, \mathcal{S}_\textrm{TE}\subset \mathcal{S}$,  $\mathcal{S}_\textrm{TR}\cap \mathcal{S}_\textrm{TE} = \varnothing$,  used for training and testing purposes, respectively. For convenience, we will refer to the cardinality of these sets using $n_\textrm{TR} = \left|\mathcal{S}_\textrm{TR}\right|$, $n_\textrm{TE} = \left|\mathcal{S}_\textrm{TE}\right|$. Concatenating the elements of $\mathcal{S}_\textrm{TR}$ horizontally, we define the matrix $S_\textrm{TR}\in\mathbb{R}^{3\times n_\textrm{tr}}$. We do not assume $S_\textrm{TR}$ is available directly. Instead, we assume the observed data are subject to additive noise
\begin{equation}
X_\textrm{TR} = S_\textrm{TR} + Y \label{eq:XTR}
\end{equation}
consistent with \eqref{eq:xmdl}. From the available $X_\textrm{TR}$ we fit a surface, estimating $\hat{\theta}$. These parameters, or more specifically $A$, define a surface ${\hat{\mathcal{S}}\subset\mathbb{R}^3}$ as an infinite collection of points indexed by coordinates $(u,v)\in\mathbb{R}$. These quantities are shown for an example surface in Figure \ref{fig:rosenbrock}.

To quantify surface fit, we consider the distance between elements $\mathbf{x}_i\in\mathcal{S}_\textrm{TE}$ and their nearest neighbor ${\mathbf{s}\left(\hat{u}_i,\hat{v}_i;A\right)\in \hat{\mathcal{S}}}$. Finding neighbor coordinates $\left(\hat{u}_i,\hat{v}_i\right)$ requires solving  \eqref{eq:argminuv}. We then define the matrix $\hat{S}_\textrm{TE}\in\mathbb{R}^{3\times n_\textrm{TE}}$ concatenating $\mathbf{s}\left(\hat{u}_i,\hat{v}_i;A\right)$ horizontally. We quantify performance with two metrics
\begin{align}
\hat{\sigma}^2_\textrm{TR} & = \frac{1}{3n_\textrm{TR}}  \left\|\hat{S}_\textrm{TR} - X_\textrm{TR}\right\|_{F}^2 \label{eq:sigmatr}\\
\hat{\sigma}^2_\textrm{TE} & = \frac{1}{3n_\textrm{TE}}  \left\|\hat{S}_\textrm{TE} - S_\textrm{TE}\right\|_{F}^2. \label{eq:sigmate}
\end{align}
Here $\hat{S}_\textrm{TR}$ are determined by $\hat{\theta}$. When $w_i=1$, \eqref{eq:sigmatr} is equivalent to \eqref{eq:sigma2}.  

Two attributes of the training data have a significant impact on algorithm performance: $n_\textrm{TR}$ and the variance of $Y$, $\sigma_Y^2$. We demonstrate these effects using a scaled-version of the Rosenbrock function (see Figure \ref{fig:rosenbrock}). We sample the reference function and apply a consistent, randomly generated rotation to all sampled points. We then partition the rotated sample points as $S_\textrm{TR}, S_\textrm{TE}$. Randomly generating i.i.d. elements $\left[Y\right]_i\sim\mathcal{N}\left(0,\sigma^2_Y\right)$, we generate $X_\mathrm{TR}$ using \eqref{eq:XTR}. From $X_\mathrm{TR}$ we estimate $\hat{\theta}$ using Algorithm \ref{alg:fitsurface}, which also determines $\hat{S}_\mathrm{TR}$ and $\hat{\sigma}^2_\textrm{TR}$. From $\hat{A}$ and $\mathcal{S}_\textrm{TE}$ we estimate $\hat{\mathcal{S}}_\textrm{TE}$ by solving \eqref{eq:argminuv} for each $\mathbf{x}_i\in\mathcal{S}_\mathrm{TE}$. Using $\hat{\mathcal{S}}_\textrm{TE}$, we determine $\hat{\sigma}^2_\textrm{TE}$ according to \eqref{eq:sigmate}. This process was repeated 100 times for multiple pairs $(n_\textrm{TR}, \sigma^2_Y)$. The averages are shown in Table \ref{table:itrslts}.

\begin{table}[htb]
	\caption{Effects of Problem Size and Noise on Results}
	\label{table:itrslts}
	\centering
	\begin{tabular}{|cc|ccccc|} 
		\hline\rule{0pt}{2.4ex}
		$n_\textrm{TR}$ & $\sigma^2_Y$ & iter. & size & $\hat{\sigma}^2_\textrm{TR}$ & $\hat{\sigma}^2_\textrm{TE}$ & ms \\[2pt]
		\hline                                                                                                                  
		50 & 2.5E-01 & 9.91 & 5.94 & 6.8E-02 & 4.9E-02 & 391.3 \\                                                               
		50 & 1.0E-02 & 10 & 14.21 & 2.5E-03 & 4.6E-03 & 404.0 \\                                                                
		50 & 2.5E-03 & 9.31 & 17.93 & 8.1E-04 & 2.4E-03 & 353.5 \\                                                              
		50 & 1.0E-04 & 5.84 & 30.28 & 2.5E-04 & 1.6E-03 & 221.4 \\                                                              
		\hline                                                                                                                  
		100 & 2.5E-01 & 10 & 6.91 & 6.8E-02 & 3.3E-02 & 449.2 \\                                                                
		100 & 1.0E-02 & 9.99 & 14.89 & 3.1E-03 & 1.7E-03 & 448.1 \\                                                             
		100 & 2.5E-03 & 9.61 & 18.86 & 9.6E-04 & 8.2E-04 & 434.8 \\                                                             
		100 & 1.0E-04 & 6.5 & 32.85 & 1.9E-04 & 3.4E-04 & 282.8 \\                                                              
		\hline                                                                                                                  
		1000 & 2.5E-01 & 10 & 12.79 & 7.0E-02 & 1.1E-02 & 1343.2 \\                                                             
		1000 & 1.0E-02 & 9.92 & 23.09 & 3.3E-03 & 3.0E-04 & 1331.9 \\                                                           
		1000 & 2.5E-03 & 7.63 & 34.79 & 8.3E-04 & 9.9E-05 & 1044.3 \\                                                           
		1000 & 1.0E-04 & 6.86 & 60.27 & 3.7E-05 & 2.7E-05 & 937.7 \\                                                            
		\hline                                                                                                                  
	\end{tabular}                                                   
\end{table}  

In Table \ref{table:itrslts}, the {\it iter.} column represents the average number of AM iterations ($t$ in Algorithm \ref{alg:fitsurface}). We limited 10 as the maximum number of iterations which we found sufficient for establishing the appropriate surface order. The {\it size} column in Table \ref{table:itrslts} indicates the number of control points in $A$: $(n_u+1)(n_v+1)$. We emphasize this is not $d$ to avoid dependence on $n_\textrm{TR}$. There are two trends to observe. First, size increases with $n_\textrm{TR}$. Second, size increases as $\sigma_y^2$ decreases. Both of these trends are due to the BIC. Similarly, we find that larger model sizes are associated with lower $\hat{\sigma}^2_\textrm{TR}$ and $\hat{\sigma}^2_\textrm{TE}$ since the BIC hedges against over-fitting. The final column depicts the average run time for surface fitting (excluding point selection). Computations were preformed in MATLAB on a Late 2016 MacBook Pro (2.9 GHz Quad-Core i7). The separable problem was parallelized among 4 workers with no GPU support.

\begin{figure}
	\centering
	\subfloat[Plane]{
		\includegraphics{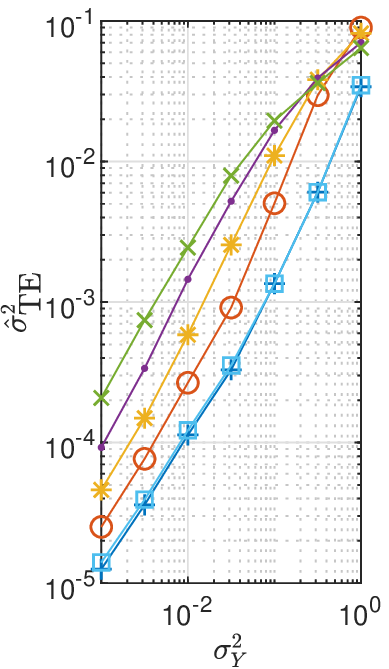}
		\label{fig:fit:plane}
	}
\subfloat[Rosenbrock function]{
	\includegraphics{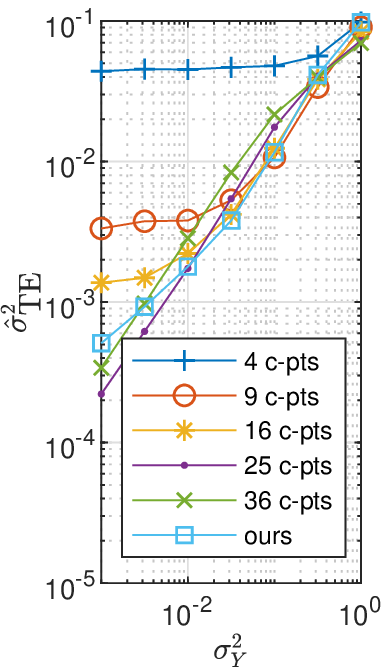}
	\label{fig:fit:rosen}
}
	\caption{ Contrasting testing error as a function of noise. For Figure \ref{fig:fit:plane} and Figure \ref{fig:fit:rosen} the latent surfaces represent a plane and Rosenbrock function, respectively. The legend indicates the number of control points assumed in fixed-order models.}\label{fig:fit_order}
\end{figure}

To further demonstrate the benefits of automatic order selection, we contrast fitting errors of fixed-order models. In addition to the Rosenbrock test surface, we now consider a plane as a second latent test surface. Intuitively we expect low-order models to perform poorly on the Rosenbrock test surface (under fitting), while high order models perform poorly on the planar surface (over fitting). This is confirmed in Figure \ref{fig:fit_order}. Additionally we find our approach performs well in both cases, automatically selecting a reasonable order for the latent surface from noisy measurements.

\begin{figure}
	\centering
	\includegraphics[width=\hsize]{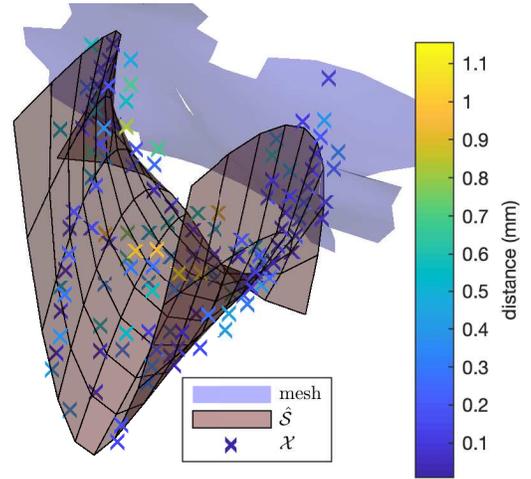}
	\caption{Local surface approximated from human procedure data. The nonparametric surface presented to human navigators is labeled {\it mesh}. Our parametric surface approximation and the intermediate point cloud are labeled $\hat{S}$ and $\mathcal{X}$, respectively. The marker colors for $\mathcal{X}$ indicates the $\ell 2$ distance (normal) to the approximated surface in mm.}\label{fig:leftatrium_mesh}
\end{figure}

To demonstrate the complete algorithm performance, we revisit the human procedure data depicted in Figure \ref{fig:leftatrium}. The occupancy map was constructed from position data after applying an unpublished algorithm \cite{Walker2016b} to reduce biological motion \cite{McClelland2013}. From the occupancy map we extract a dense binary matrix $V$ indicating interior volume with cubic voxels with 1mm edge lengths. We identify $\mathcal{X}$ applying Algorithm \ref{alg:pointselect} to a cubic volume with edge lengths 15mm. This collection was indicated as the point cloud in Figure \ref{fig:leftatrium}. Using Algorithm \ref{alg:fitsurface} we fit a surface $\hat{\mathcal{S}}$ to $\mathcal{X}$. Results are shown in Figure \ref{fig:leftatrium_mesh}. In this case $n_x = 132$, and we used $w_i=1$. For the fitted surface, $n_u = 1$, $n_v=3$, and $\hat{\sigma}^2 \approx 0.05$. This case presents a region of the anatomy with high curvature, yet the resulting error is well below quantization of the data. To contrast our results with the surface presented to the physician navigators, we limit {\it display} to only mesh vertices representing a nearest point to one of the elements of $\mathcal{X}$. The median distance from $\mathcal{X}$ to the nonparametric display surface (nearest vertex) is 3.38 mm in contrast to 0.24mm (orthogonal distance) to our parametric surface.

\section{Conclusions}\label{sec:Conclusion}
We have demonstrated an iterative algorithm for surface approximation that converges quickly and is robust against over fitting. We feel this approach is well suited for the unique challenges associated with autonomous navigation of catheters for cardiac ablation surgery.

Our algorithm for surface fitting only address the local search problem. In other words we assume the initial parameter selections, $\mathbf{u}^{(0)}$,$\mathbf{v}^{(0)}$, are near the global minimum. We select $\mathbf{u}^{(0)}$, $\mathbf{v}^{(0)}$ by projection $\mathcal{X}$ onto a 2-dimensional subspace (SVD). The usefulness of this strategy will depend on the curvature of the latent surface. For example, this is not reasonable when the point cloud approximates a cylinder. Limiting the extent of the surface approximation is one mitigation strategy. Larger surfaces may require global search or stitching together multiple surfaces approximations and presents an opportunity for further research.

\appendix
\subsection{Gradient and Hessian of the Separable Problem}\label{app:gradhess}
The Jacobian of \eqref{eq:skron} can be expressed
\begin{equation}
\mathbf{J}_\mathbf{s} = \begin{bmatrix}
\mathbf{b}(v)\otimes \mathbf{b}'(u)  & 
\mathbf{b}'(v)\otimes \mathbf{b}(u) 
\end{bmatrix}^T \bar{A}.
\end{equation}
Here we use $\mathbf{b}'(u)$ to represent the derivative of \eqref{eq:bbold} with respect to $u$. Subsequently we will use $\mathbf{b}''(u)$ to represent the second derivative. Their analytic derivation is straight forward from \eqref{eq:bernsteinpoly}. The gradient of \eqref{eq:g} is then
\begin{equation}
\nabla g(u,v;\mathbf{x},A) = -\mathbf{J}_\mathbf{s}\left(\mathbf{x} - \mathbf{s}(u,v;A)\right)\label{eq:ggrad}
\end{equation}

For the convenience, we define the auxiliary scalar values
\begin{align}
c_u &= \left(
\mathbf{b}(v)\otimes \mathbf{b}''(u)\right)^T \bar{A} \left(\mathbf{x} - \mathbf{s}(u,v;A)\right)\\
c_v &= \left(
\mathbf{b}''(v)\otimes \mathbf{b}(u)\right)^T \bar{A} \left(\mathbf{x} - \mathbf{s}(u,v;A)\right)\\
c_{uv} &= \left(
\mathbf{b}'(v)\otimes \mathbf{b}'(u)\right)^T \bar{A} \left(\mathbf{x} - \mathbf{s}(u,v;A)\right).
\end{align}
The Hessian of \eqref{eq:g} is then
\begin{equation}
\nabla^2 g(u,v;\mathbf{x}, A) =\\ \mathbf{J}_\mathbf{s}\, \mathbf{J}_\mathbf{s}^T - \begin{bmatrix}c_u & c_{uv} \\ c_{uv} & c_{vv}\end{bmatrix}.
\end{equation}


\section*{Acknowledgment}
We are grateful to Dr. Burkhard H{\"u}gl for providing procedure data facilitating this research. Nathan Kastelein, Paul F. Rebillot III and \.{I}lker Tunay contributed ideas and helpful comments on the manuscript.

\bibliographystyle{ieeetr}
\bibliography{bib,medical_bib}

\end{document}